\documentclass[10pt,twocolumn,letterpaper]{article}

\usepackage{cvpr}
\usepackage{times}
\usepackage{epsfig}
\usepackage{graphicx}
\usepackage{amsmath}
\usepackage{amssymb}
\usepackage{subfig}

\usepackage[breaklinks=true,bookmarks=false]{hyperref}

\cvprfinalcopy % *** Uncomment this line for the final submission

\def\cvprPaperID{****} % *** Enter the CVPR Paper ID here

\begin{document}

%%%%%%%%% TITLE
\title{Object Detection using Domain Randomization and Generative Adversarial Refinement of Synthetic Images}

\author{Fernando Camaro Nogues\qquad Andrew Huie\qquad Sakyasingha Dasgupta\\
Ascent Robotics, Inc. Japan \\
{\tt\small \{fernando, andrew, sakya\}@ascent.ai}
}

\maketitle
%\thispagestyle{empty}

%%%%%%%%% ABSTRACT
\begin{abstract}
   In this work, we present an application of domain randomization and generative adversarial networks (GAN) to train a near real-time object detector for industrial electric parts, entirely in a simulated environment. Large scale availability of labelled real world data is typically rare and difficult to obtain in many industrial settings. As such here, only a few hundred of unlabelled real images are used to train a Cyclic-GAN network, in combination with various degree of domain randomization procedures. We demonstrate that this enables robust translation of synthetic images to the real world domain.  We show that a combination of the original synthetic (simulation) and GAN translated images, when used for training a Mask-RCNN object detection network achieves greater than 0.95 mean average precision in detecting and classifying a collection of industrial electric parts. We evaluate the performance across different combinations of training data.
\end{abstract}

%%%%%%%%% BODY TEXT
\section{Introduction}

The successful examples of deep learning require a large number of manually annotated data, which can be prohibitive for most applications, even if they start from a pre-trained model in another domain and only require a fine-tuning phase in the target domain. \newline
An effective way to eliminate the cost of the expensive annotation is to train the model within a simulated environment where the annotations can be also automatically generated. However, the problem with this approach is that the generated samples (in our case images) may not follow the same distribution as the real domain, resulting in what is known as the reality-gap. Several approaches exist that try to reduce this apparent gap. One such method is domain randomization (\cite{domain_rand}, \cite{sadeghi2017}). In this,  several rendering parameters of the scene can be randomized, like the color of objects, textures, lights, etc, thus effectively enabling the model to see a very wide distribution during training, and seeing the real distribution as one variation in it. \newline
Another approach that directly tries to minimize this reality-gap is to refine the synthetic images so that they look more realistic. One possible way to build such a refiner is by using a generative adversarial training framework \cite{Shrivastava2017}. An alternative and more indirect approach to reduce the negative effect of this reality-gap is to use again the GAN framework, but in this case, directly on the features of some of the last layers of the network being trained for the specific target task \cite{Ganin2016}. \\
In this work we present an experimental use case of an object detector in a real industrial application setting, which is trained with different combinations of synthetic images and refined synthetic images (synthetic images refined to look more realistic). We evaluate our method robustly across various combinations of training data.

\section{Synthetic Image Generation with Domain Randomization}
The architecture to produce the synthetic images for our experiments is composed of two main parts. First, the physics simulation engine, Bullet\footnote{https://pybullet.org/wordpress/} is used to place the objects in a physically consistent configuration after letting them fall from a random position. Second, the ray tracing rendering library POV-Ray\footnote{http://www.povray.org/} is used to render an image based on this configuration. In POV-Ray we introduce domain randomization, by randomizing several parameters, namely, the number of lights and their color, the color and texture of each part of the target objects and the scene floor plane, as well as the camera position. The camera position is drawn from a uniform distribution in a rectangular prism that is 10cm above the floor plane, with a squared base of side 20cm and 10cm height. Although the location of the camera was uniform, the camera was always pointing to the global coordinates origin with no rolling angle. This variation of the camera position was intended to achieve robustness against different positions of the camera in the real world.

\section{Refinement of synthetic images by adversarial training}
An alternative way we consider to reduce the reality-gap is to use the GAN framework to refine the synthetic images  to look more realistic. Here, we selected the Cyclic-GAN \cite{cyclicgan} architecture since it only requires two sets of unpaired examples, one for each domain, the synthetic and the real one. The original synthetic images of size 1024x768 were too large for the training of our Cyclic-GAN model, as such,  instead of resizing the image, we opted for training on random crops of size 256x256. This way we can train in the original pixel density and exploit the fact that our generators are fully convolutional networks, such that during the inference phase we can still input the original full-size image. 
\begin{figure}[h]
\begin{center}
\includegraphics[width=0.45\linewidth]{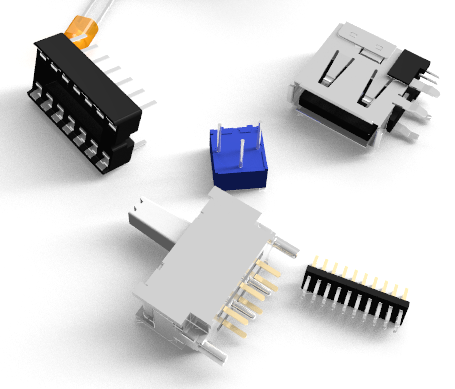}
\includegraphics[width=0.45\linewidth]{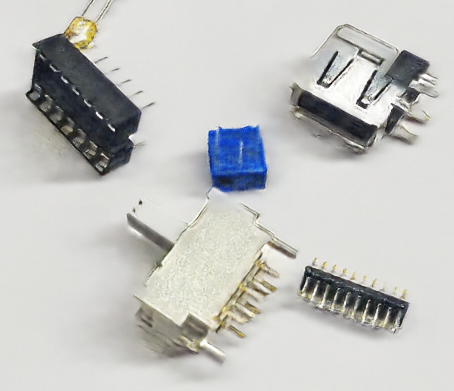}
\end{center}
   \caption{Left: example of synthetic image. Right: corresponding synthetic image after translation to real domain. The USB socket has gained a more realistic reflection, and the switch has gained a realistic surface texture and color.}
\label{fig:sim2real}
\end{figure}\\
We notice that after training, one particular target object lost its color and turned gray, while the remaining objects were refined in a realistic manner without loosing their original color. We think that this was mainly due to the particular architecture of the discriminators. The discriminator model final layer consisted of a spatial grid of discriminator neurons whose receptive field with respect to the input image was too small to capture that object. In order to solve this we added more convolutional layers to the discriminator models. This effectively increased the receptive field size. Furthermore,  instead of substituting one grid of discriminators by another, we preferred to maintain both, one with small receptive field intended to discriminate details of the objects, and another with large receptive field, that can understand the objects as a whole (Fig. \ref{fig:2disclayers} in Appendix). The final loss was computed as the mean of all individual discriminator units for both of these two layers. This small modification enabled us to maintain the color of all the objects. The Cyclic-GAN model was trained using 10K synthetic images and 256 real images. Fig.\ref{fig:sim2real} shows an example of the resulting image with our model that translates from synthetic domain to the real domain; see Fig. \ref{fig:moresim2real} in Appendix for more examples.

%-------------------------------------------------------------------------
\section{Experiments}
In this section we compare different combinations of training data and its impact on the mAP for object detection with a Mask-RCNN model \cite{maskrcnn}. As a test dataset we have used 100 real images. \\
The different types of datasets used for training were:
$S_{fix}$ : synthetic images with fixed object colors without texture, and white background.
$S_{fix\ \rightarrow real}$: translated images from $S_{fix}$ to the real domain.
$S_{rand-tex}$: synthetic images with objects and background with randomized colors but without texture.
$S_{rand+tex}$: synthetic images with objects and background with randomized colors and texture.
See Fig. \ref{fig:training_arch} in the appendix for a general overview of the training architecture and Fig. \ref{fig:typesimages} for some examples of different types of images employed. \\
The target objects to be detected, consisted of 12 tiny electronic parts for which accurate 3D CAD models were available (Fig. \ref{fig:parts}). In all the experiments we used 10K training samples, the same number of training iterations and the same hyperparameters.

The object detection performance for the different combinations of datasets used in the experiments are presented in Table  \ref{table:results} . Using a training set made purely of one type of data resulted in a mAP  below 0.9 in most cases, with the exception of the case with $S_{rand+ tex}$. Overall, the best detection results were obtained when the refined synthetic images set ($S_{fix\rightarrow real}$) was combined with high variation randomized data ($S_{rand+tex}$). The results indicate that neither domain randomization or GAN based refinement is enough on its own to get sufficient performance. In combination, they reduce the reality-gap effectively, resulting in a significant boost in performance (\textit{see the real-time object detection video at \url{https://youtu.be/Q-WeXSSnZ0U}}). Refer to Fig. \ref{fig:training_curves} for the training curves associated with the different experiments, and to Fig. \ref{fig:detection_result} for some detection result images.

\begin{table}[!th]
\begin{center}
\begin{tabular}{|l|c|}
\hline
Training data & mAP (0.5 IoU)\\
\hline\hline
100\% $S_{fix}$ & 0.812\\ % experiment 10
100\% $S_{fix\rightarrow real}$ & 0.874\\ % experiment 11 
100\% $S_{rand-tex}$ & 0.867 \\ % experiment 6
100\% $S_{rand+tex}$ & 0.911 \\ % experiment 7

20\% $S_{fix}$ and 80\% $S_{rand+tex}$  & 0.914\\ % experiment 9 
20\% $S_{fix\rightarrow real}$ and 80\% $S_{rand+tex}$  & 0.955\\ % experiment 5 
50\% $S_{fix\rightarrow real}$ and 50\% $S_{rand+tex}$  & 0.950\\ % experiment 4 
\hline
\end{tabular}
\end{center}
\caption{Performance of the Mask-RCNN network for the different training datasets.}
\label{table:results}
\end{table}

{\small
\bibliographystyle{ieee}
\bibliography{egbib}

\section*{Appendix}
In Fig. \ref{fig:training_arch} we provide a schematic overview of the object detection training data generation pipeline.
\begin{figure}[ht]
\begin{center}
\includegraphics[keepaspectratio=true,scale=0.7]{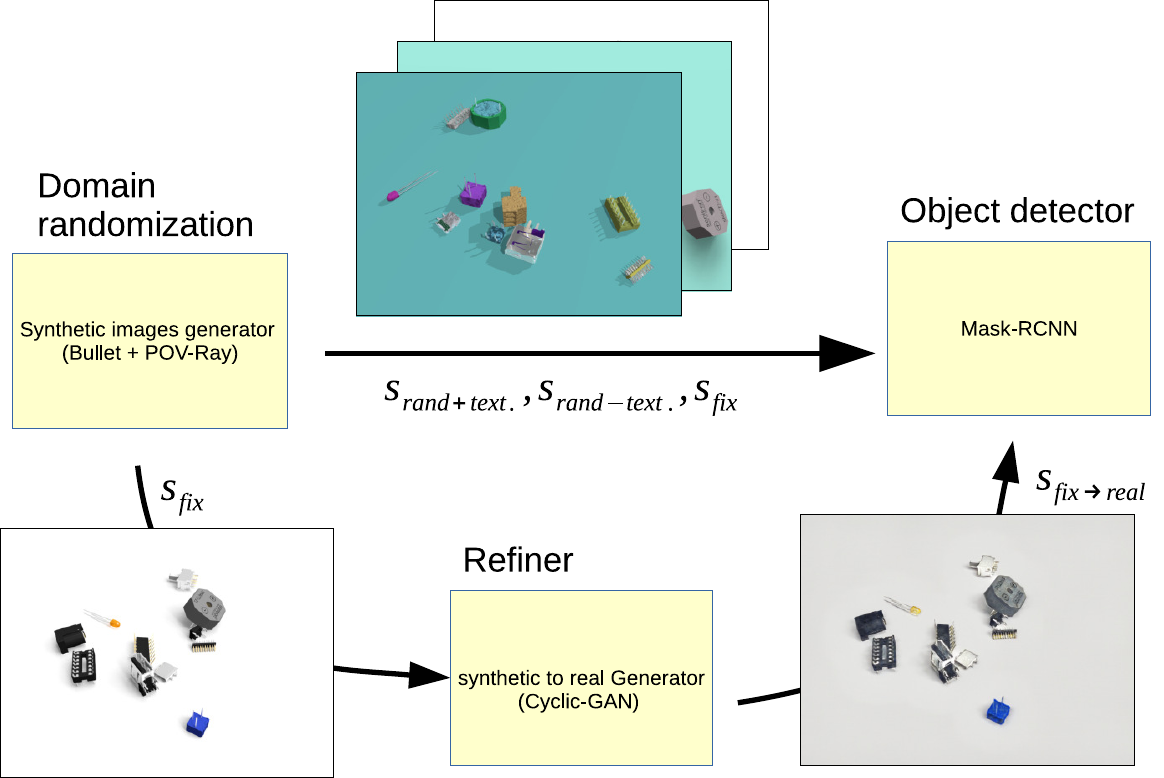}
\end{center}  
\caption{General architecture for training the object detector.}
\label{fig:training_arch}
\end{figure}

\begin{figure}[h]
\begin{center}
\includegraphics[width=1\linewidth]{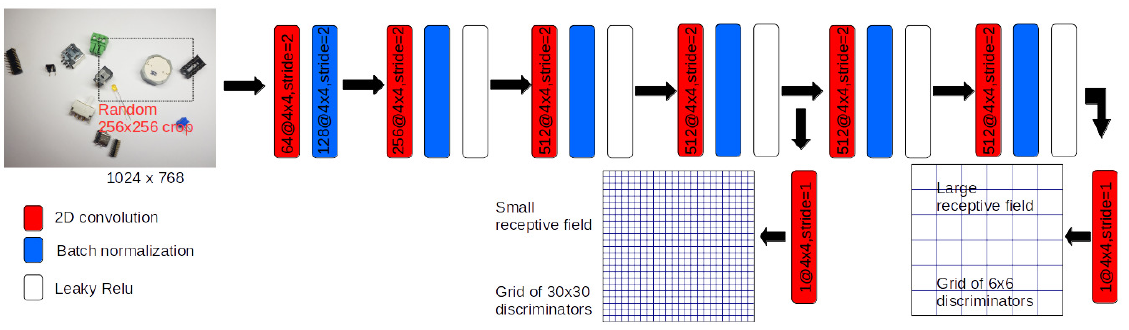}
\end{center}
   \caption{Discriminator network with two grid layers of discriminator cells, one with small receptive field and the other with bigger receptive field.}
\label{fig:2disclayers}
\end{figure}

\begin{figure*}
\begin{center}
\begin{tabular}{cc}
\includegraphics[width = 0.4\linewidth]{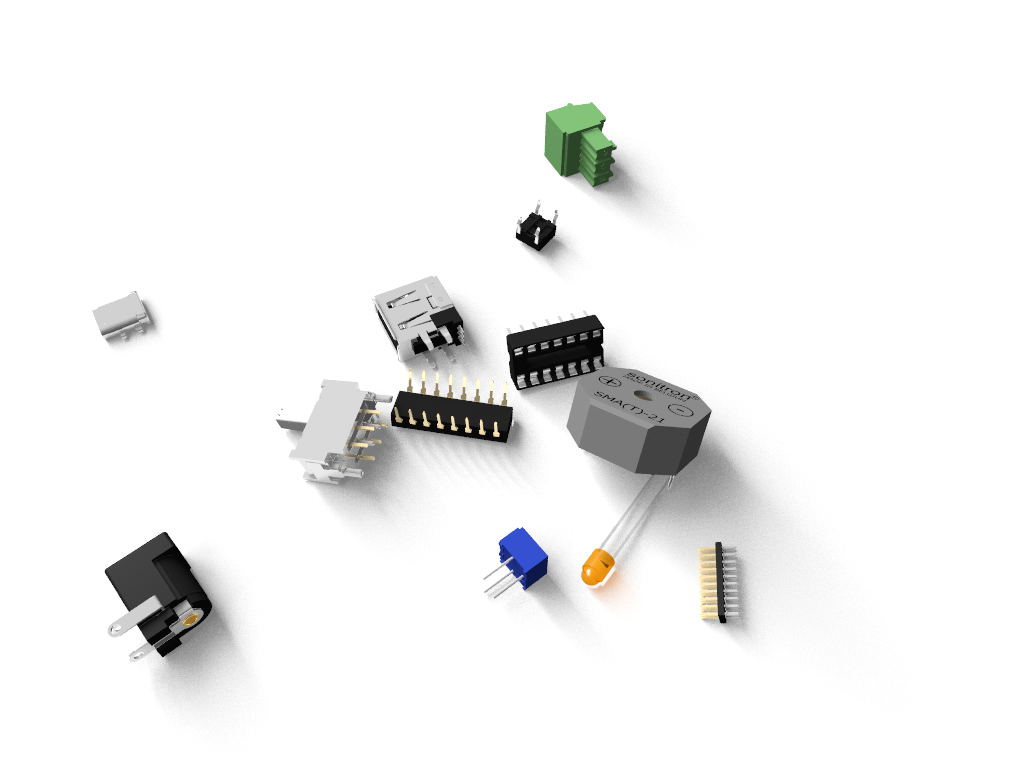} &
\includegraphics[width = 0.4\linewidth]{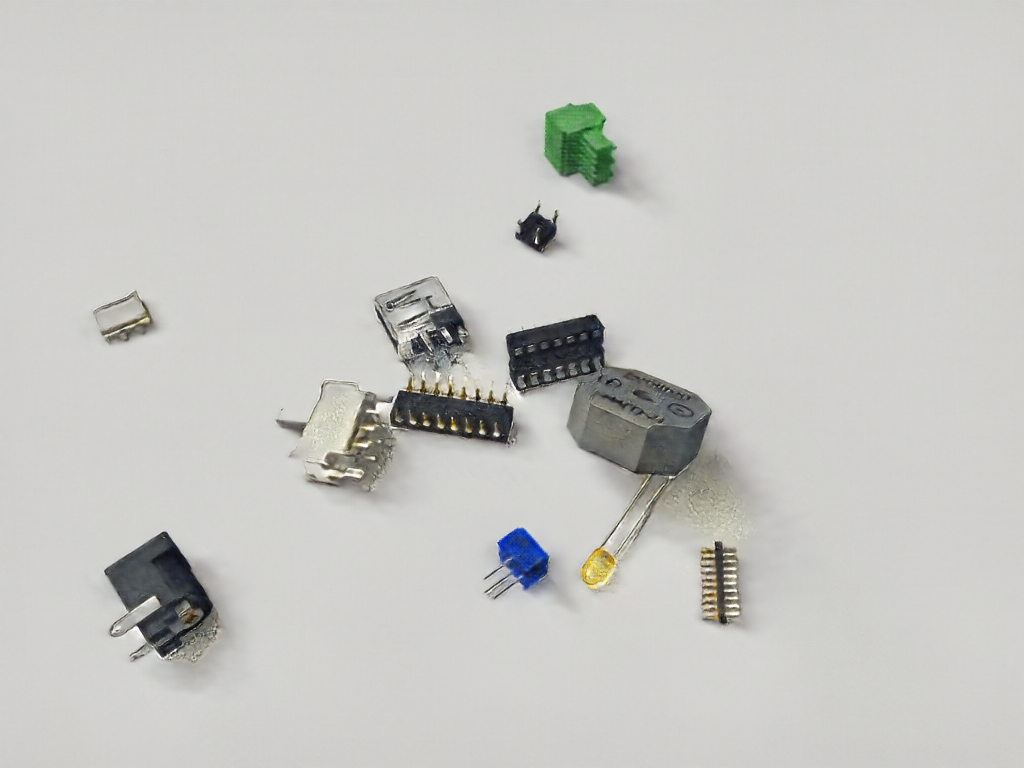} \\
\includegraphics[width = 0.4\linewidth]{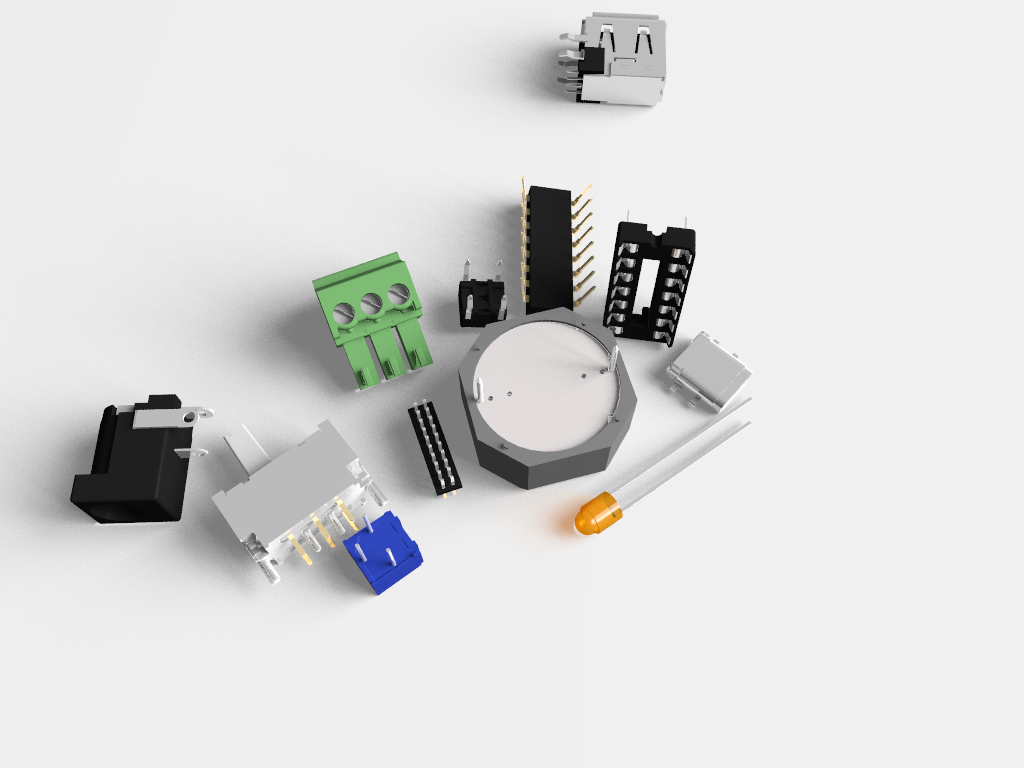} &
\includegraphics[width = 0.4\linewidth]{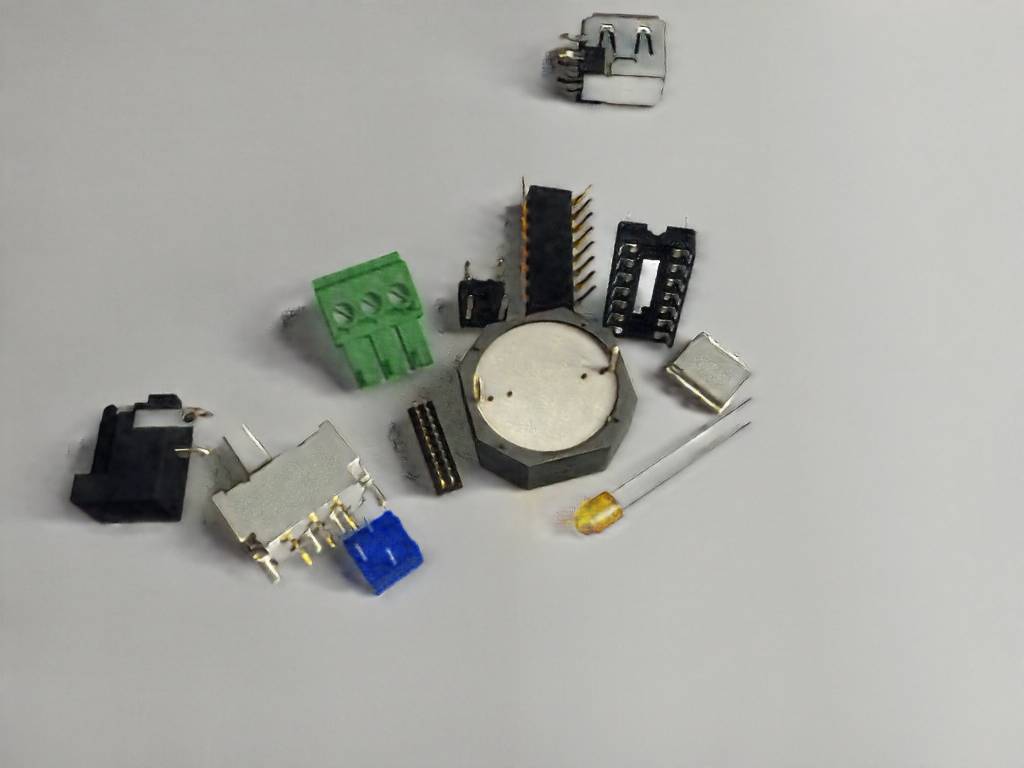}\\
\includegraphics[width = 0.4\linewidth]{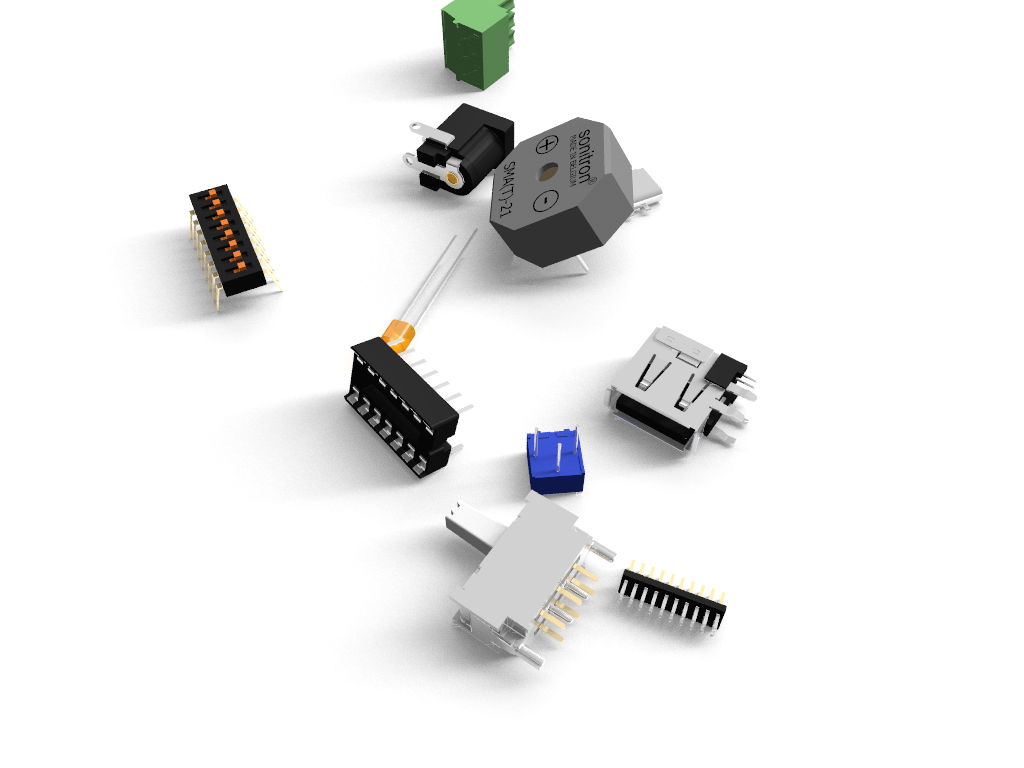} &
\includegraphics[width = 0.4\linewidth]{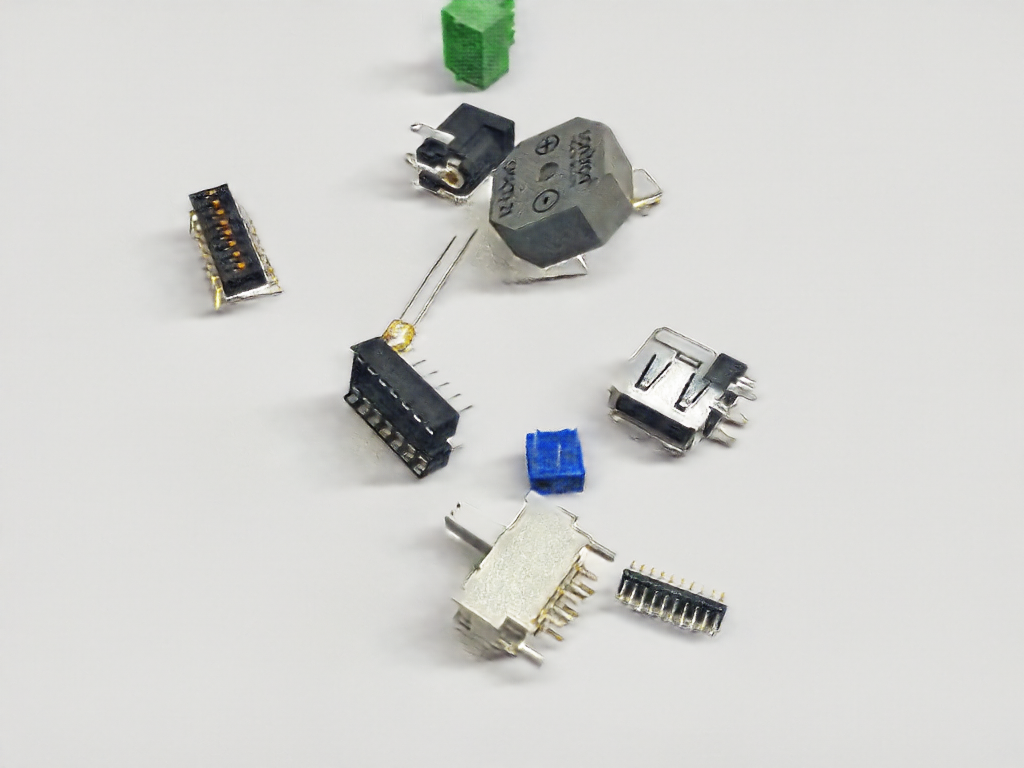}  \\
\includegraphics[width = 0.4\linewidth]{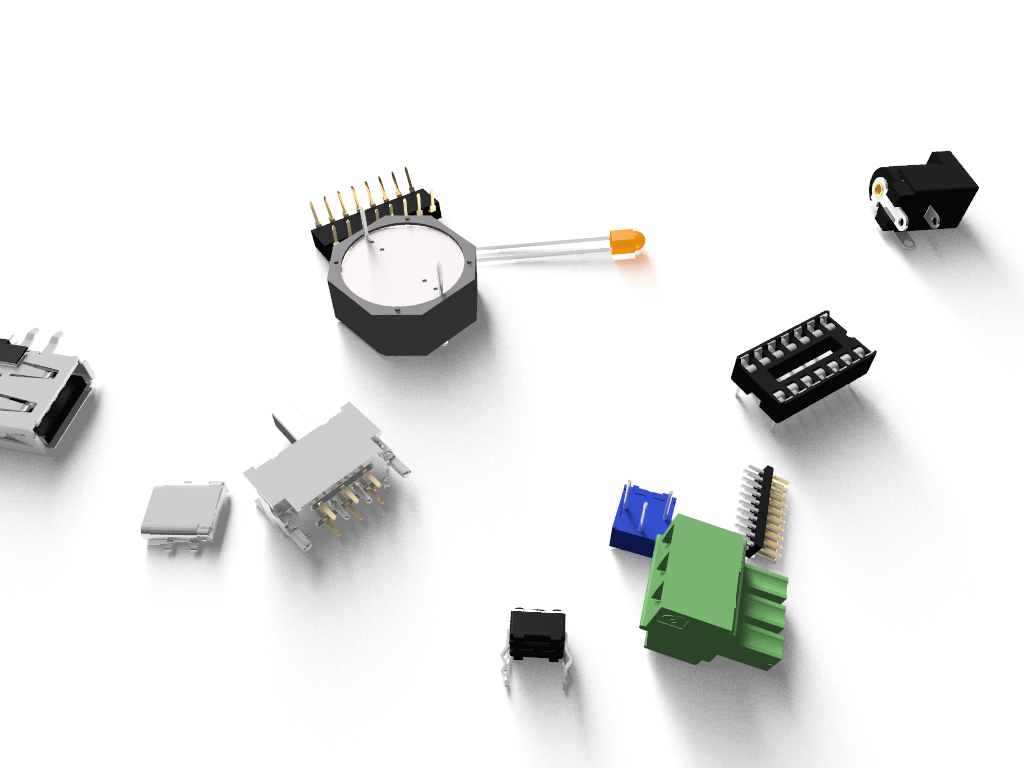} &
\includegraphics[width = 0.4\linewidth]{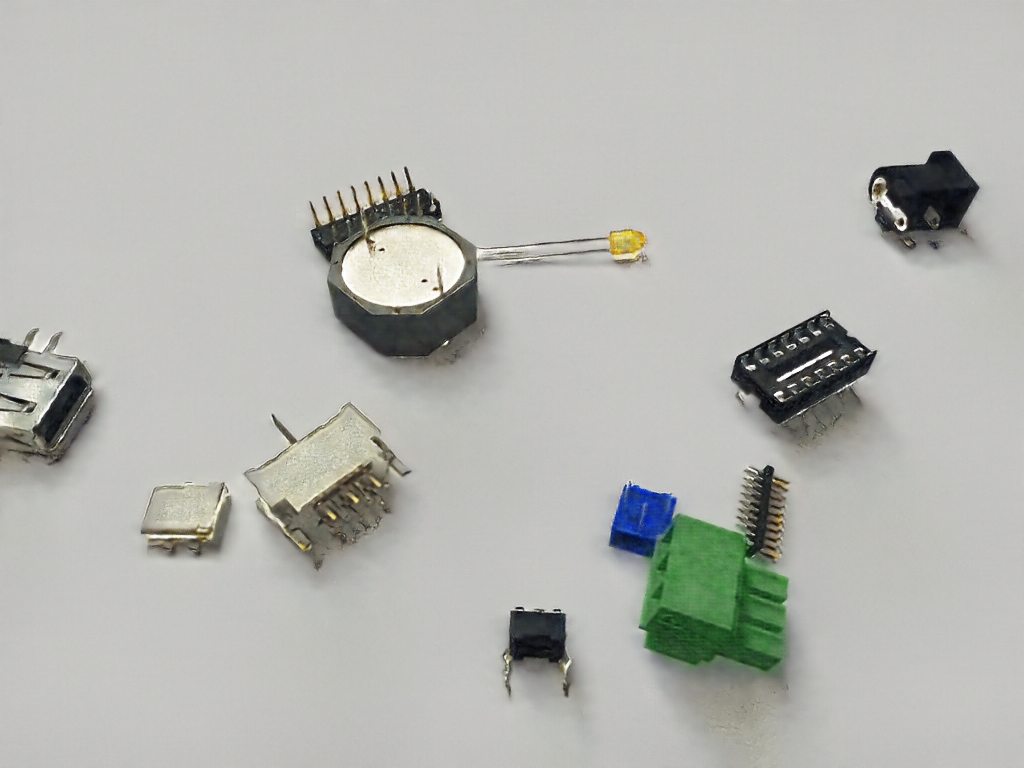} \\
\end{tabular}
\end{center}
\caption{Left column: images from $S_{fix}$. Right column: corresponding refined images ($S_{fix\rightarrow real}$).}
\label{fig:moresim2real}
\end{figure*}

\begin{figure*}
\begin{center}
\begin{tabular}{cc}
\subfloat[Example of $S_{fix}$ image]{\includegraphics[width = 0.5\linewidth]{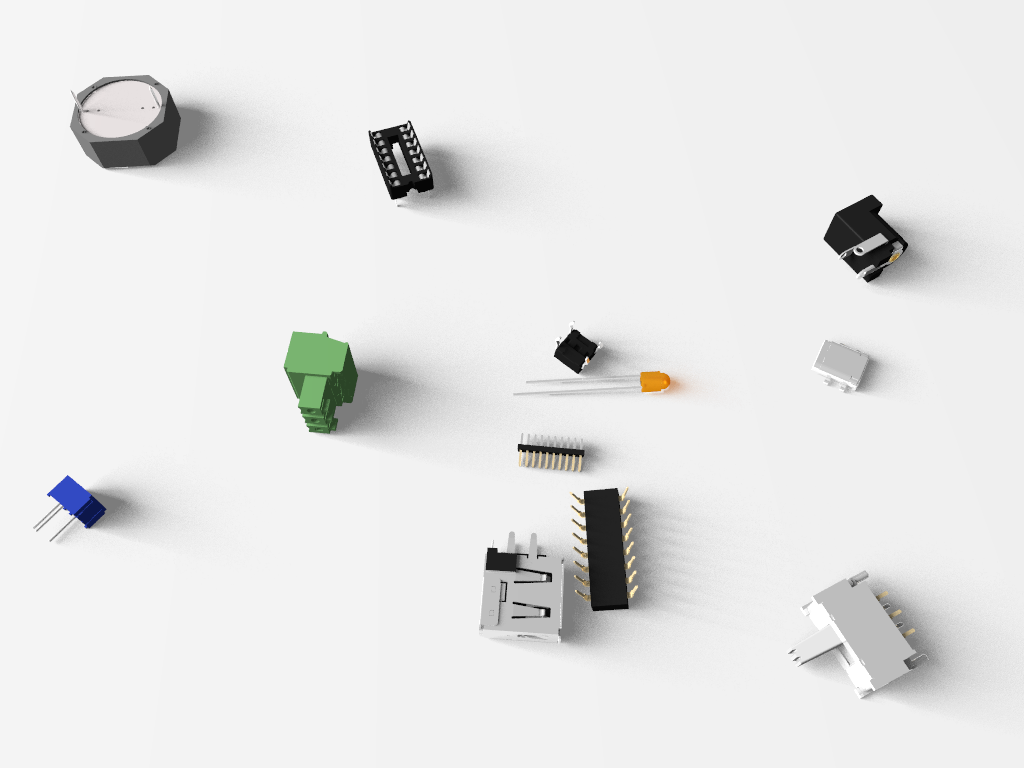}} &
\subfloat[Example of $S_{rand-tex}$ image]{\includegraphics[width = 0.5\linewidth]{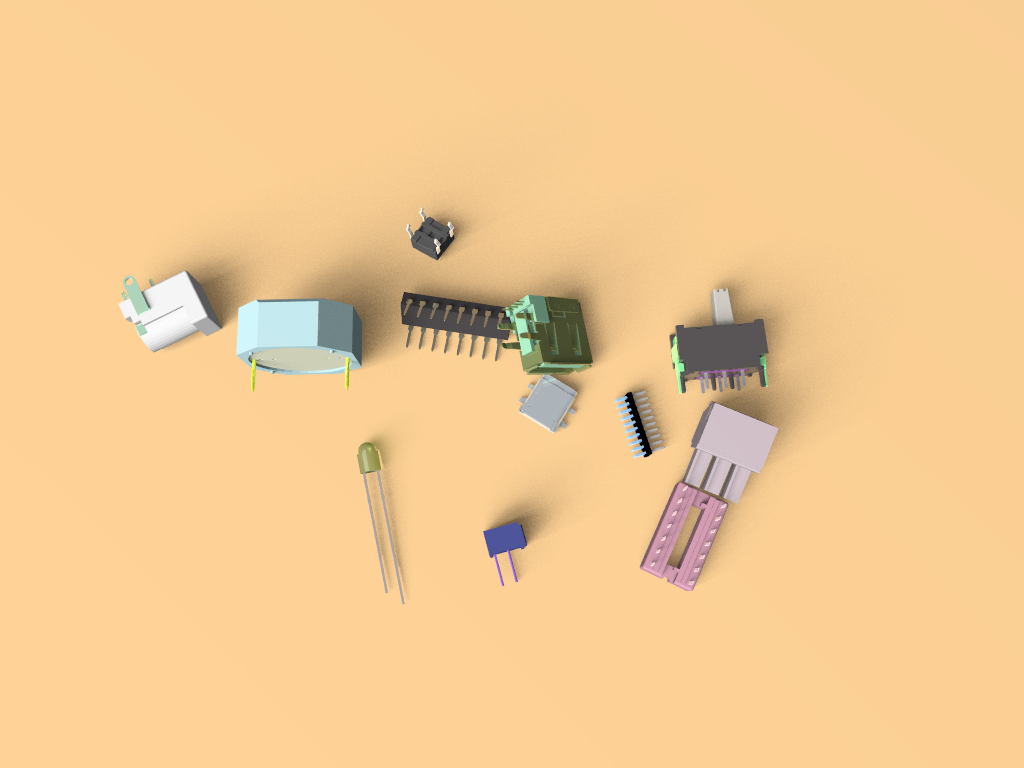}} \\
\subfloat[Example of $S_{rand+tex}$ image]{\includegraphics[width = 0.5\linewidth]{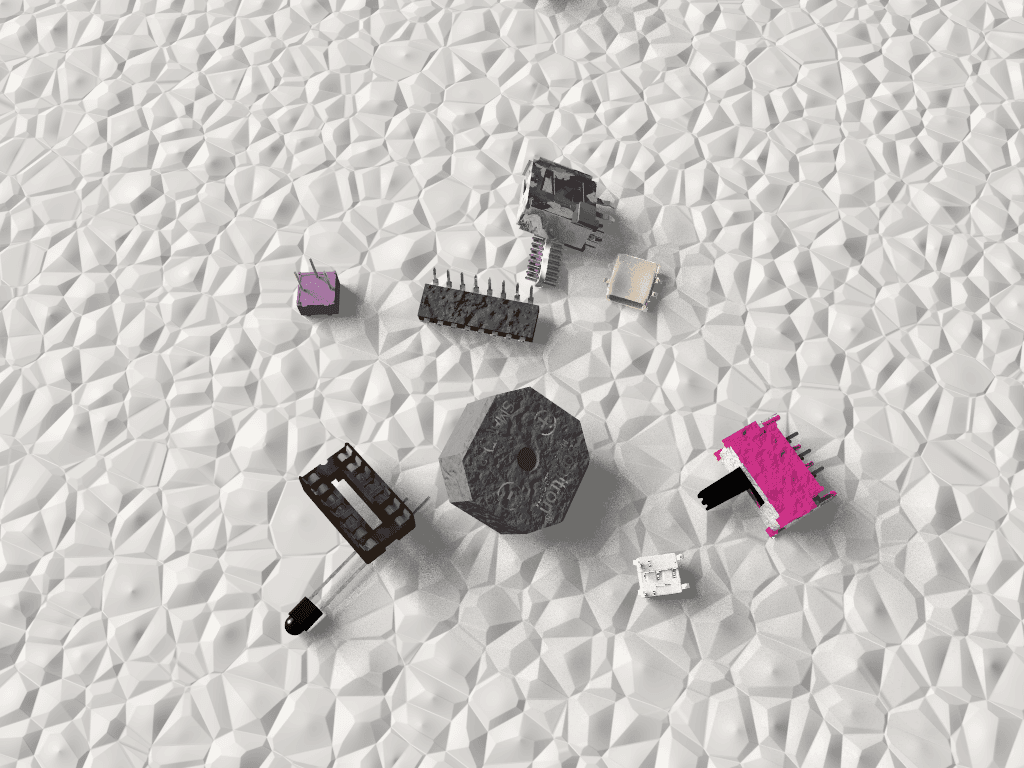}} &
\subfloat[Example of a real image used to train the cyclic-GAN]{\includegraphics[width = 0.5\linewidth]{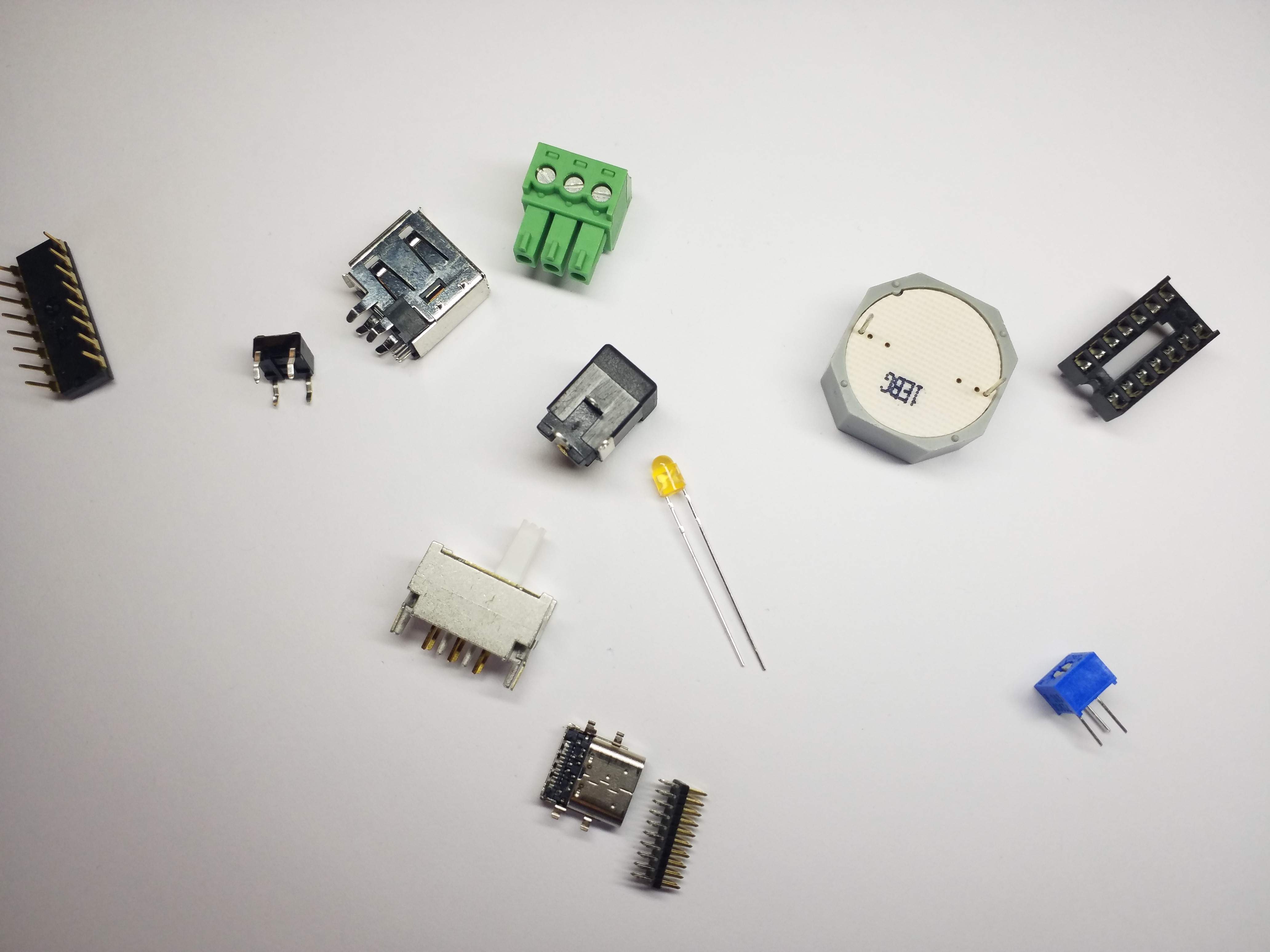}} \\  
\end{tabular}
\end{center}
\caption{Examples of different types of images employed in the experiments.}
\label{fig:typesimages}
\end{figure*}

\begin{figure*}
\begin{center}
\begin{tabular}{ccc}
\subfloat[tactile switch]{\includegraphics[width = 0.2\linewidth]{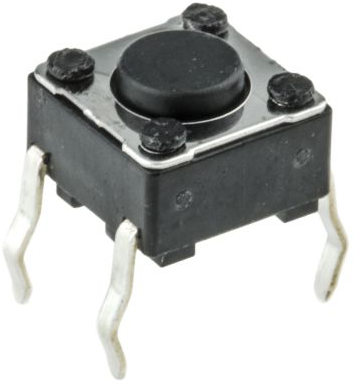}} &
\subfloat[pin header]{\includegraphics[width = 0.2\linewidth]{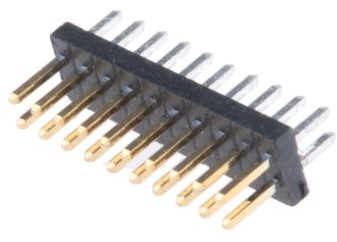}} & 
\subfloat[3 way cable mount screw terminal]{\includegraphics[width = 0.2\linewidth]{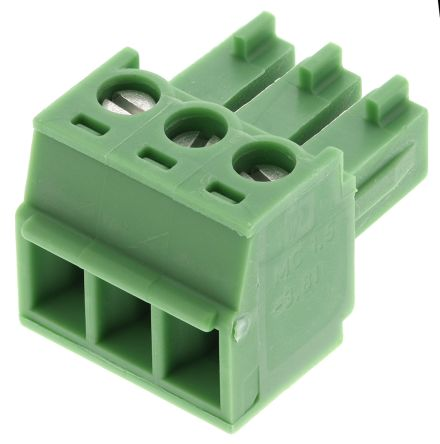}} \\
\subfloat[DC power jack]{\includegraphics[width = 0.2\linewidth]{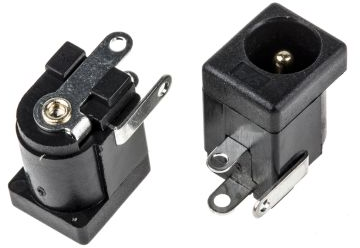}} &
\subfloat[DIP switch]{\includegraphics[width = 0.2\linewidth]{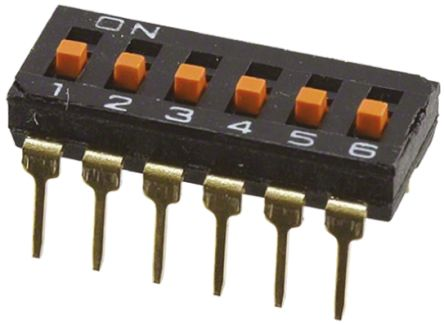}} & 
\subfloat[slide switch]{\includegraphics[width = 0.2\linewidth]{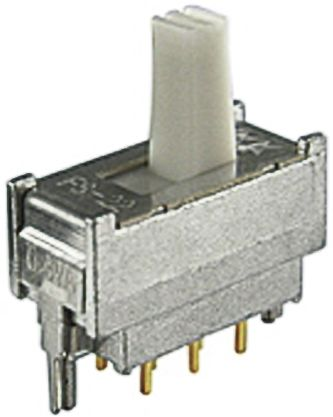}} \\
\subfloat[led]{\includegraphics[width = 0.2\linewidth]{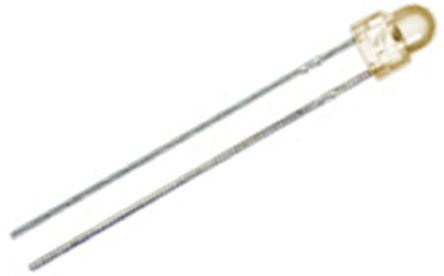}} &
\subfloat[IC socket]{\includegraphics[width = 0.2\linewidth]{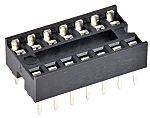}} & 
\subfloat[trimmer]{\includegraphics[width = 0.2\linewidth]{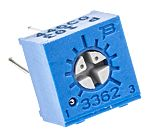}} \\
\subfloat[buzzer]{\includegraphics[width = 0.2\linewidth]{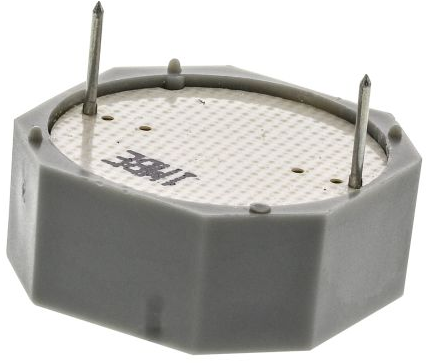}} &
\subfloat[USB type A socket]{\includegraphics[width = 0.2\linewidth]{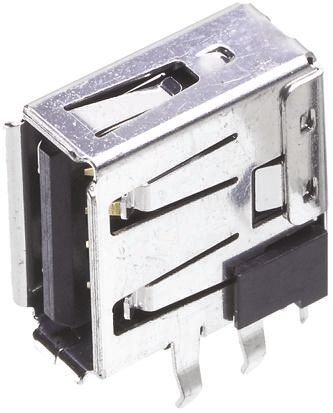}} & 
\subfloat[USB type C socket]{\includegraphics[width = 0.2\linewidth]{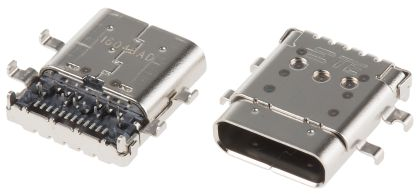}} \\
\end{tabular}
\end{center}
\caption{Electronic parts used in the experiments.}
\label{fig:parts}
\end{figure*}

\begin{figure*}[h]
\begin{center}
\includegraphics[width=0.8\linewidth]{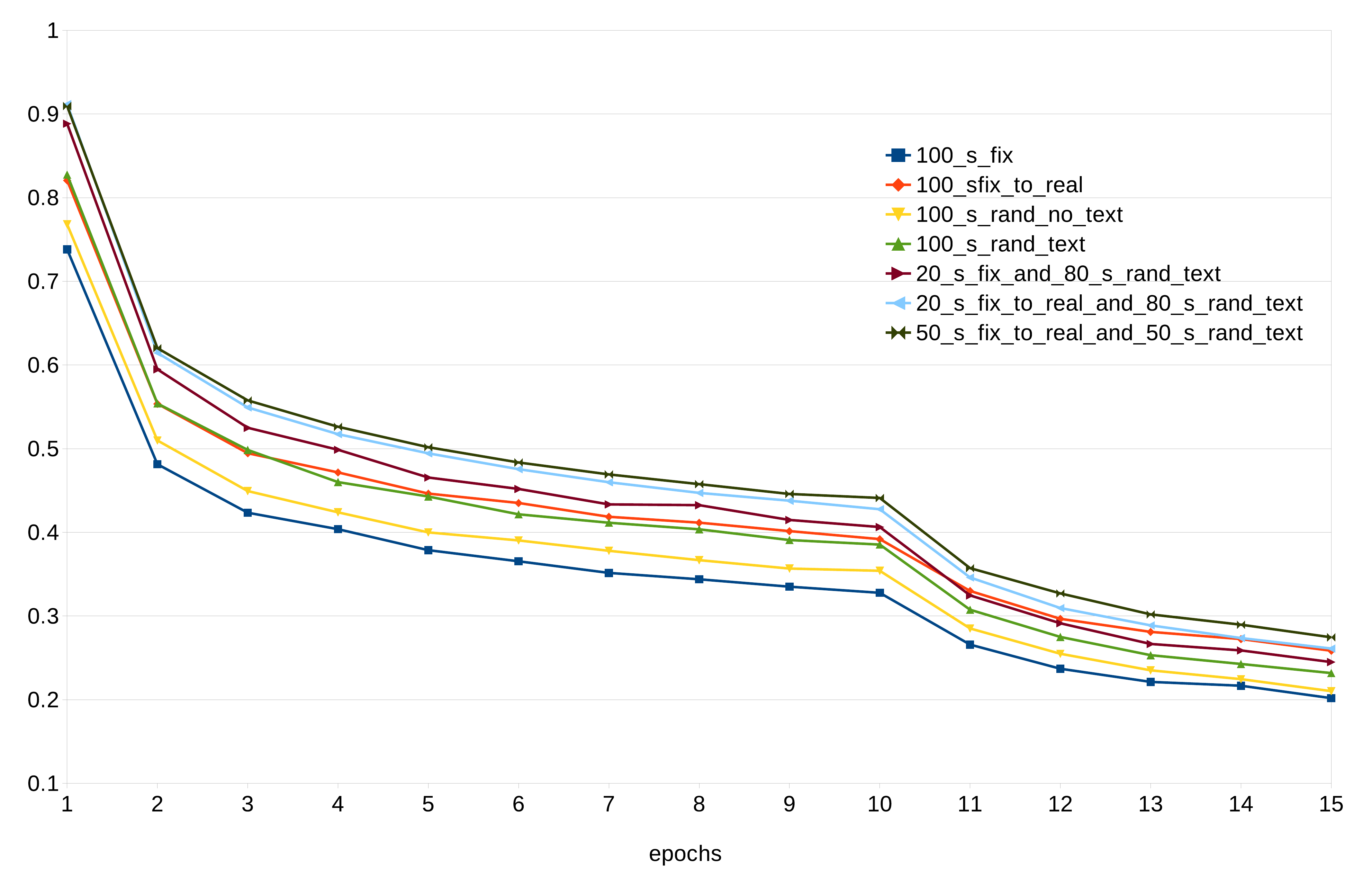}
\end{center}
   \caption{Mask-RCNN training loss. The model was trained by fine-tuning a Mask-RCNN model pre-trained on the COCO dataset. First by training only the mask-rcnn heads (without training the region proposal network or the backbone model) for 10 epochs with a learning rate of 0.002, and then the whole network for another 5 epochs with a learning rate of 0.0002. We used a SGD optimizer with a momentum of 0.9. The configurations that achieved the better performances, "20\% $S_{fix\rightarrow real}$ and 80\% $S_{rand+tex}$" and "50\% $S_{fix\rightarrow real}$ and 50\% $S_{rand+tex}$", are the ones that had worse loss values during training. We think that this is because these datasets were more difficult, but at the end prepared the model better for the also difficult real test dataset.}
\label{fig:training_curves}
\end{figure*}

\begin{figure*}[h]
\begin{center}
\includegraphics[width=0.8\linewidth]{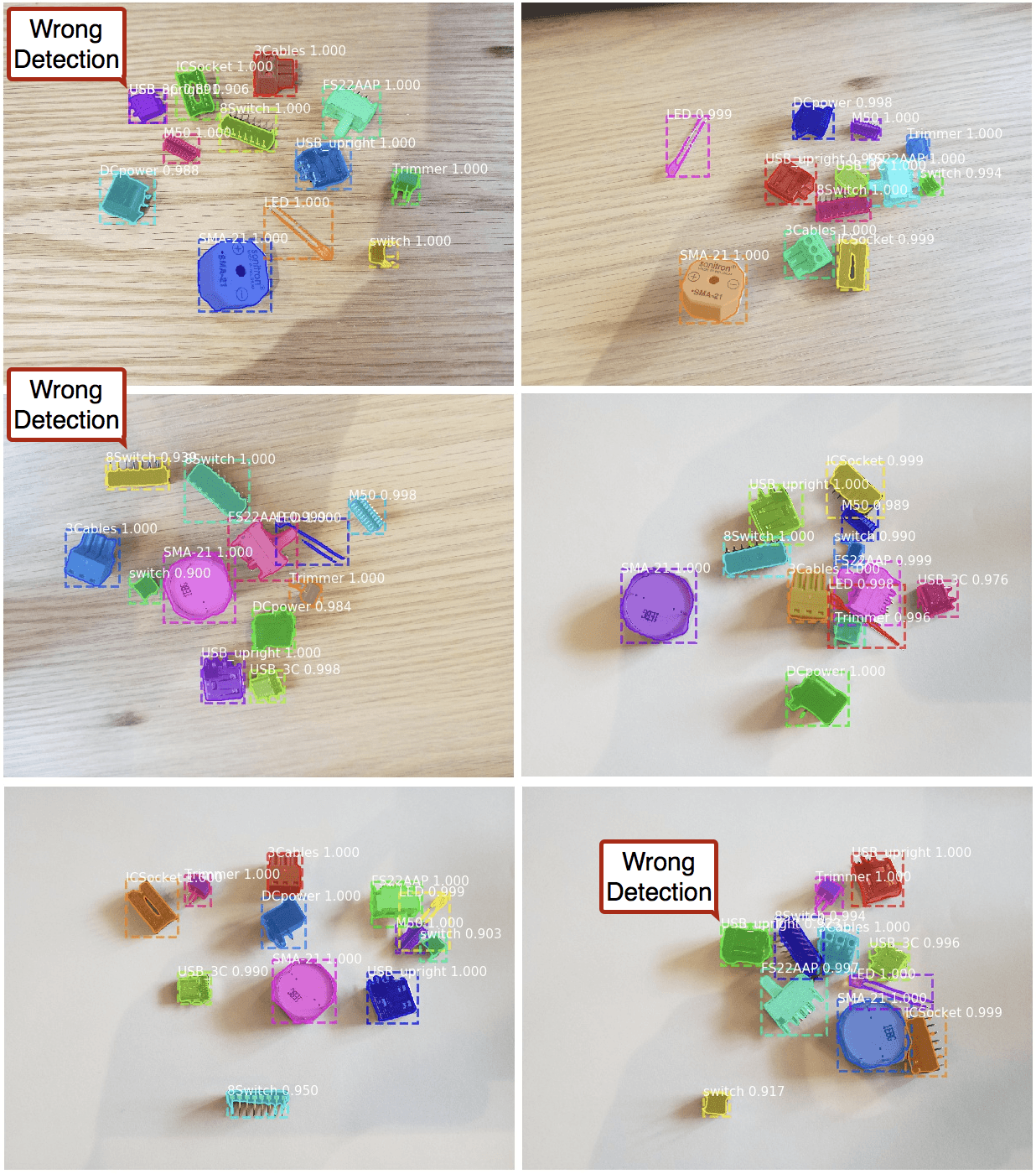}
\end{center}
   \caption{Example of detection results for 20\% $S_{fix\rightarrow real}$ and 80\% $S_{rand+tex}$.}
\label{fig:detection_result}
\end{figure*}

}

\end{document}